\documentclass{article}

\usepackage[preprint]{neurips_data_2024}

\usepackage[utf8]{inputenc} % allow utf-8 input
\usepackage[T1]{fontenc}    % use 8-bit T1 fonts
\usepackage{url}            % simple URL typesetting
\usepackage{booktabs}       % professional-quality tables
\usepackage{amsfonts}       % blackboard math symbols
\usepackage{nicefrac}       % compact symbols for 1/2, etc.
\usepackage{microtype}      % microtypography
\usepackage{xcolor}         % colors

\usepackage{microtype}
\usepackage{graphicx}

\usepackage{textcomp}
\usepackage{mdframed}
\usepackage{amsmath,amssymb} % define this before the line numbering.
\usepackage{color, colortbl}
\usepackage{enumitem}
\usepackage{multirow}
\usepackage{bbm}
\usepackage{wrapfig}

\usepackage{amsmath,amsfonts,bm}

\def\eqref#1{equation~\ref{#1}}
\def\1{\bm{1}}

\DeclareMathAlphabet{\mathsfit}{\encodingdefault}{\sfdefault}{m}{sl}
\SetMathAlphabet{\mathsfit}{bold}{\encodingdefault}{\sfdefault}{bx}{n}

\usepackage{tikz}
\usepackage{comment}

\usepackage{mdframed}
\newcommand{\sectioncolor}{violet}
\usepackage{algorithm}
\usepackage{algorithmic}

\usepackage[font=small]{caption}
\usepackage{subcaption}

\usepackage{xspace}
\newcommand{\multimed}{{\fontfamily{lmtt}\selectfont{\textsc{MultiMed}}}\xspace}

\definecolor{Gray}{gray}{0.9}
\definecolor{citecolor}{HTML}{2980b9}
\definecolor{linkcolor}{HTML}{c0392b}

\usepackage[pagebackref=true,breaklinks=true,colorlinks,bookmarks=false,citecolor=blue,linkcolor=red]{hyperref}

\usepackage{cleveref}
\crefformat{section}{Sec.~#2#1#3}
\crefformat{subsection}{Sec.~#2#1#3}
\crefformat{subsubsection}{Sec.~#2#1#3}
\crefformat{figure}{Fig.~#2#1#3}
\crefformat{equation}{Eq.~#2#1#3}
\crefformat{table}{Tab.~#2#1#3}
\crefformat{algorithm}{Alg.~#2#1#3}

\title{\multimed: Massively Multimodal and Multitask Medical Understanding}

\author{%
  Shentong Mo\\
  Carnegie Mellon University\\
  \texttt{shentongmo@gmail.com}
  \And
  Paul Pu Liang\\
  Massachusetts Institute of Technology\\
  \texttt{ppliang@mit.edu}
}

\begin{document}

\maketitle

\begin{abstract}

Biomedical data is inherently multimodal, consisting of electronic health records, medical imaging, digital pathology, genome sequencing, wearable sensors, and more. The application of artificial intelligence tools to these multifaceted sensing technologies has the potential to revolutionize the prognosis, diagnosis, and management of human health and disease. However, current approaches to biomedical AI typically only train and evaluate with one or a small set of medical modalities and tasks. This limitation hampers the development of comprehensive tools that can leverage the rich interconnected information across many heterogeneous biomedical sensors.
To address this challenge, we present \multimed, a benchmark designed to evaluate and enable large-scale learning across a wide spectrum of medical modalities and tasks.
\multimed\ consists of 2.56 million samples across ten medical modalities such as medical reports, pathology, genomics, and protein data, and is structured into eleven challenging tasks, including disease prognosis, protein structure prediction, and medical question answering. Using \multimed, we conduct comprehensive experiments benchmarking state-of-the-art unimodal, multimodal, and multitask models.
Our analysis highlights the advantages of training large-scale medical models across many related modalities and tasks.
Moreover, \multimed\ enables studies of generalization across related medical concepts, robustness to real-world noisy data and distribution shifts, and novel modality combinations to improve prediction performance. \multimed\ will be publicly available and regularly updated and welcomes inputs from the community.
\end{abstract}

\vspace{-2mm}
\section{Introduction}\label{sec:intro}
\vspace{-2mm}

The integration of artificial intelligence in medicine has opened avenues for diagnostics and treatment planning~\cite{Abramson2024alphafold3,Jumper2021alphafold,saab2024capabilities,yang2024advancing}. Medical data is inherently multimodal, consisting of electronic health records, medical imaging, digital pathology, genome sequencing, wearable sensors, and more~\cite{acosta2022multimodal,liang2022highmmt,liang2022foundations,stahlschmidt2022multimodal,tu2024towards}. However, most advances in biomedical AI typically only train and evaluate with one or a small set of medical modalities and tasks~\cite{Kenneth2011ppmi,Morozov2020MosMedDataCC,Nguyen2022vindr,shih2019rsna,wang2020covidx,zhang2020clinically}. For example, while there has been substantial progress in medical image analysis~\cite{braintumor,shih2019rsna,wang2020covidx}, these models do not also incorporate data from genomics~\cite{chang2013tcga,Papatheodorou2017atlas}, proteins~\cite{Berman2000pdb}, digital pathology~\cite{chen2020pathomic,jaume2023modeling}, EEGs~\cite{gu2021eeg}, or wearable and ambient sensors~\cite{liang2021learning,majumder2017wearable,mo2023multiiot,mo2024IoTLM} that can help monitor patient health on a day-to-day basis beyond rare imaging appointments~\cite{heikenfeld2018wearable,pantelopoulos2009survey}. Each medical modality can offer unique and synergistic information towards understanding patient conditions and outcomes, and can substantially increase the volume and variety of information available for holistic analysis~\cite{liang2022foundations,liang2023quantifying}. Due to a lack of large-scale, centralized resources that represent the full breadth of biomedical knowledge~\cite{acosta2022multimodal,stahlschmidt2022multimodal,tu2024towards}, it is difficult to build comprehensive machine learning technologies that leverage the rich interconnected information across modalities and tasks.

\begin{figure}[t]
\centering
\vspace{-0mm}
\includegraphics[width=0.9\linewidth]{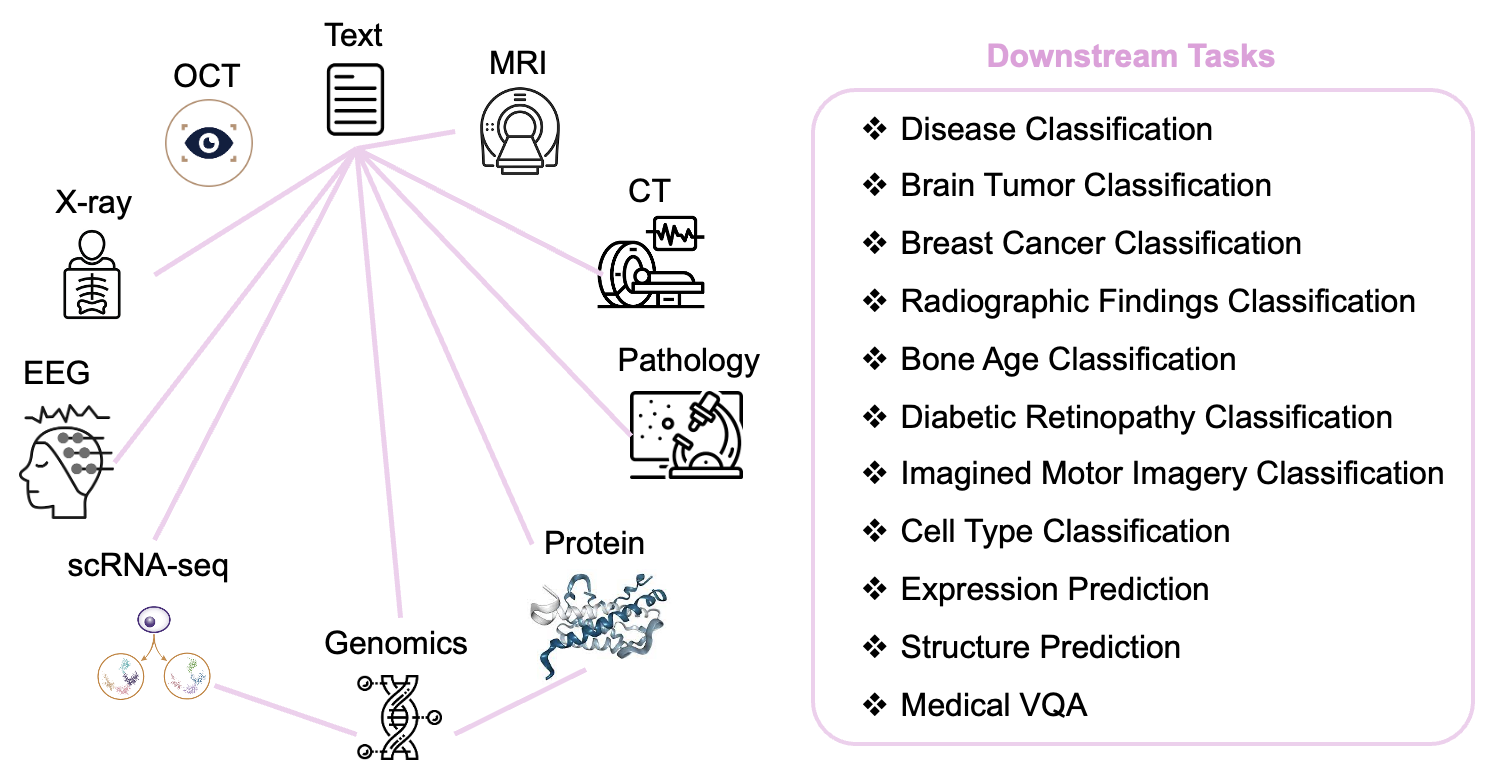}
\vspace{-0mm}
\caption{\multimed\ is a large-scale benchmark for representation learning in the medical domain, consisting of 2.56M samples, 10 rich modalities, and 11 challenging tasks in real-world medical scenarios. We also present new challenges for impactful applications involving text, OCT, X-ray, CT, MRI, Pathology, EEG, genomics, scRNA-seq, and proteins. The lines represent modality pairings present in individual \multimed\ datasets, such as those between text and MRI, text and pathology, as well as text, genomics, and proteins among others.}
\vspace{-2mm}
\label{fig: title_img}
\end{figure}

To bridge this gap, we introduce \multimed, a new benchmark designed specifically for multimodal and multitask medical data analysis. 
\multimed\ offers 2.56 million samples encompassing ten diverse modalities such as medical reports, pathology, genomics, and protein data, and is structured into eleven challenging tasks, including disease prognosis, protein expressions, and medical question answering (see Figure~\ref{fig: title_img} for a summary).
This benchmark not only facilitates the development of large-scale multimodal and multitask medical models but also provides the opportunity to study crucial societal impacts of data bias, robustness, and generalization.
\multimed\ sets a new standard for evaluating and developing new multimodal biomedical AI technologies, which can have a great impact on medical diagnosis, personalized medicine, and improving patient health outcomes~\cite{acosta2022multimodal,stahlschmidt2022multimodal}.

Overall, our contributions can be summarized into four main folds:
\begin{enumerate}[noitemsep,topsep=0pt,nosep,leftmargin=*,parsep=0pt,partopsep=0pt]
    \item \multimed\ is a large-scale benchmark featuring 2.56 million samples across ten rich modalities including text, imaging (OCT, X-ray, CT, MRI), electrophysiology (EEG), and molecular data (genomics, scRNA-seq, protein). \multimed\ also includes paired modalities, such as text and MRI, text and pathology, as well as text, genomics, and proteins to study multimodal data integration.
    \item \multimed\ encompasses eleven challenging medical tasks, such as disease classification, brain tumor classification, and medical VQA, providing a broad testing ground across various medical diagnosis and prediction problems.
    \item We propose and benchmark multimodal and multitask learning models that integrate data from multiple sources while performing several tasks simultaneously. On \multimed\ tasks, these models significantly push forward the state-of-the-art in medical data analysis.
    \item \multimed\ also enables studies of generalization to new medical modalities and tasks, positive transfer across related concepts, robustness to real-world noise and distribution shifts, and novel modality combinations which offer insights into these models' capabilities, limitations, and practical applicability.
\end{enumerate}

\vspace{-2mm}
\section{Related Work}
\vspace{-2mm}

We cover related work in biomedical artificial intelligence, multimodal machine learning, and similar benchmarks for machine learning in healthcare.

\noindent\textbf{Multimodal learning benchmarks.}
There has been significant progress in multimodal learning benchmarks~\cite{agrawal2017vqa,lee2023holistic,liang2021multibench,zadeh2018multimodal} that has highlighted the importance of effectively combining information from different sensory channels to improve the accuracy and robustness of predictive models~\cite{baltruvsaitis2018multimodal,liang2022foundations}.
Inspired by these developments, \multimed\ extends these principles into the medical domain, addressing the unique challenges posed by medical data modalities and tasks~\cite{acosta2022multimodal,stahlschmidt2022multimodal,tu2024towards}.
By incorporating complex and diverse data types such as imaging, genomic, and electronic health records, \multimed\ aims to catalyze similar advancements in medical data analysis, leveraging the rich potential of multimodal integration to enhance diagnostic and prognostic capabilities.

\noindent\textbf{Unimodal medical benchmarks.}
Unimodal benchmarks have been pivotal in advancing specific areas of medical research~\cite{Kenneth2011ppmi,Morozov2020MosMedDataCC,Nguyen2022vindr,shih2019rsna,wang2020covidx,zhang2020clinically}.
For example, benchmarks such as BRATS~\cite{menze2015brats} for brain tumor segmentation using MRI scans and ISIC~\cite{Codella2018isic} for skin lesion analysis using dermatologic images have significantly advanced the state-of-the-art in their respective fields. 
These benchmarks focus on refining model performance on single data types, aiming to improve accuracy and efficiency in narrow, well-defined problem spaces.
However, unimodal benchmarks cannot handle the interconnectedness of medical modalities in different real-world scenarios~\cite{acosta2022multimodal,stahlschmidt2022multimodal,tu2024towards}. \multimed\ differentiates itself by focusing on cross-modal integration and multitask learning, yielding a more holistic approach for medical data analysis.

\noindent\textbf{Multimodal medical benchmarks.}
Multimodal learning for healthcare has gained attention due to its impact on assisting doctors in the diagnosis process through diverse medical signals~\cite{acosta2022multimodal,chen2017multimodal,jaume2023modeling,lipkova2022artificial}. Benchmarks like the Medical Segmentation Decathlon~\cite{Antonelli2022medical} and MIMIC-III~\cite{Johnson2016MIMIC-III} have focused on integrating different imaging modalities or combining imaging data with electronic health records.
However, most existing multimodal benchmarks do not encompass the breadth or depth offered by \multimed. 
For instance, integrating genomic data, proteomics, and high-dimensional modalities like scRNA-seq is often overlooked. 
\multimed\ enables large-scale multimodal integration from diverse medical signals and the investigation of novel modality combinations.

\vspace{-2mm}
\section{\multimed: A Massively Multimodal and Multitask Medical Benchmark}
\vspace{-2mm}

In this section, we first provide details about our \multimed\ benchmark, designed to promote the development of machine learning models capable of handling complex, multimodal medical datasets. 
Our benchmark is constructed to focus on three dimensions of diversity: organ and cell type, modality, and task. Each dimension is crafted to ensure that the benchmark covers a broad spectrum of medical data types and challenges.

\subsection{Organ \& cell type diversity}

\multimed\ first includes data related to multiple organ systems and cell types, including brain, breast, bone, eye, and other critical areas often examined in medical diagnostics. 
This variety allows for the exploration of disease patterns across different body systems.
In addition to organs, \multimed\ also extends to cellular-level data, featuring modalities like genomics and scRNA-seq.
By integrating organ and cell type diversity, \multimed\ can help discover the molecular and cellular mechanisms underlying various medical conditions.

\subsection{Modality diversity}

To understand these underlying organs and cells, \multimed\ includes an extensive array of medical sensing modalities. These include:
\begin{enumerate}[noitemsep,topsep=0pt,nosep,leftmargin=*,parsep=0pt,partopsep=0pt]
    \item Imaging Modalities~\cite{wang2020covidx,zhang2020clinically,braintumor,Kermany2018LargeDO,Lhncbc}: Optical Coherence Tomography (OCT), X-ray, CT, MRI, and pathology images. These modalities provide spatial resolutions ranging from macroscopic organ structures to microscopic cellular details.
    We comprise 84,495 OCT images, 194,922 X-ray images, 617,775 CT scans, 7,023 MRI scans, and 27,560 pathology images.
    \item Electrophysiological Data~\cite{Aristimunha2023moabb}: EEG data offers insights into the electrical activity of the brain. \multimed\ consists of 120,000 samples designed for the classification of imagined motor imagery time-series data.
    \item Molecular Data~\cite{chang2013tcga,Papatheodorou2017atlas,Berman2000pdb}: Genomic, scRNA-seq, and protein data each provides a different perspective of biological status at the molecular level. We include 12,560 samples of genomic sequences and 270,000 samples of scRNA-seq data to support expression prediction at the single-cell level. We also include a total of 131,487 protein sequences for protein structure prediction.
    \item Text~\cite{ncbi}: Clinical notes that complement raw medical signals with rich, descriptive medical narratives with one million image-text pairs.
\end{enumerate}
The diversity of modalities in \multimed\ challenges current models to handle heterogeneous data types and integrate potentially synergistic information present in various medical modalities.

\subsection{Task diversity}

Task diversity is another cornerstone of the \multimed\ benchmark, which enables us to test the adaptability of learning models to multiple eleven medical tasks each with their unique challenges:
\begin{enumerate}[noitemsep,topsep=0pt,nosep,leftmargin=*,parsep=0pt,partopsep=0pt]
    \item Disease classification: This task involves categorizing patient data into disease categories based on symptoms, laboratory results, and imaging data. It tests the model's ability to recognize and differentiate between a wide range of diseases from common ailments to rare conditions.
    \item Brain tumor classification: Models are trained to identify and classify various types of brain tumors using MRI scans. This requires precise imaging analysis capabilities to distinguish between tumor types, which often appear similar to non-specialist algorithms.
    \item Breast cancer classification: Utilizing mammography and histopathology images, this task focuses on identifying and classifying stages of breast cancer. The challenge lies in the subtle variations between stages and the high accuracy required for clinical applicability.
    \item Radiographic findings classification: This task involves classifying findings in X-ray and CT images, such as fractures or lung nodules. The complexity arises from the diverse range of possible findings and their presentations in images.
    \item Bone age classification: Based on hand X-rays, this task estimates the skeletal maturity of a patient, which is crucial for diagnosing growth disorders in pediatrics. The models must be precise as the implications of the results can affect treatment plans.
    \item Diabetic retinopathy classification: Models classify the severity of diabetic retinopathy by analyzing retinal photographs. The grading scale's subtlety and the disease's progressive nature make this a challenging task.
    \item Imagined motor imagery classification: Using EEG data, this task classifies the type of motor imagery a subject is thinking about, which has applications in brain-computer interfaces. The challenge is the interpretation of noisy EEG signals and their low spatial resolution.
    \item Cell type classification: From single-cell RNA sequencing data, this task involves identifying cell types based on their gene expression profiles. It requires handling high-dimensional data and distinguishing between closely related cell types.
    \item Expression prediction: Predicting the expression level of genes from various inputs such as genetic markers or environmental conditions. This task tests models' ability to handle large, sparse datasets and to model complex genomic sequences.
    \item Protein structure prediction: In this task, models predict the three-dimensional structures of proteins from their amino acid sequences. It requires significant computational power and precise modeling techniques to accurately predict structures and understand protein function.
    \item Medical visual question answering involves answering clinical questions based on medical images, requiring a deep understanding of visual content, medical knowledge, and language understanding.
\end{enumerate}
Through these diverse tasks, each with its unique challenges, \multimed\ facilitates a thorough evaluation of model performance and enables the training of more generalist biomedical AI systems.

\vspace{-2mm}
\section{Medical AI Methods Benchmarked in \multimed}
\vspace{-2mm}

In this section, we discuss a range of medical AI methods that we benchmark on \multimed. These methods include those trained on a single modality and task, as well as multimodal and multitask models. We briefly review these methods below.

\subsection{Notations}

Let $X = \{X^{(1)}, X^{(2)}, \ldots, X^{(M)}\}$ represent the set of input data where $X^{(m)}$ corresponds to the $m$-th modality. Each modality contains $N$ samples, and each sample can be represented as $x_i^{(m)}$, where $i$ indexes the sample. Similarly, let $Y = \{y_1, y_2, \ldots, y_N\}$ denote the set of labels or outputs associated with these samples, which may vary based on the specific task $T$ being performed. Tasks are defined by a function $f : X \rightarrow Y$ that models can learn to approximate.

\subsection{Unimodal single-task and multitask learning}

In traditional unimodal learning, models are trained on data from a single modality. For example, a model might be trained exclusively on MRI images or genomic data. This approach limits the ability of the model to leverage complementary information from other data types. Unimodal multitask learning extends this by allowing the model to learn from one type of data while simultaneously performing multiple tasks. For instance, a model might classify tumor types and predict treatment outcomes from the same set of pathology slides. The mathematical formulation for unimodal multitask learning can be expressed as follows:
\begin{equation}
    \theta^* = \arg\min_\theta \sum_{t=1}^T \mathcal{L}_t(f(x^{(m)};\theta), y_t),
\end{equation}
where $\theta$ represents the parameters of the model, $\mathcal{L}_t$ is the loss function for task $t$, and $y_t$ is the label for task $t$.

\subsection{Multimodal fusion methods}

To overcome the limitations of unimodal approaches, multimodal techniques integrate data from multiple modalities, aiming to exploit the complementary information available. We refer the reader to~\cite{liang2022foundations} for a comprehensive review of various methods for multimodal fusion and representation learning, but summarize several baselines that we benchmark on \multimed\ below:

\noindent\textbf{Early fusion}: Data from different modalities are combined at the input level, allowing the model to learn directly from multimodal input data. This approach is straightforward but may not handle modality-specific features effectively, and tends to require larger multimodal models due to larger input dimensionality. Early fusion can be mathematically represented as:
\begin{equation}
    x_{\text{early}} = \phi(\{x^{(1)}, x^{(2)}, \ldots, x^{(M)}\}),
\end{equation}
where $\phi$ is a fusion function, such as concatenation or summation, applied across modal inputs.

\noindent\textbf{Intermediate fusion}: Features from each modality are extracted separately and then combined at one or more hidden layers within the model. This allows the model to process each modality according to their unique information before integration. The function for intermediate fusion looks like:
\begin{equation}
    h_{\text{inter}} = \psi(\{h^{(1)}, h^{(2)}, \ldots, h^{(M)}\}),
\end{equation}
where $g^{(m)}$ is a function that extracts features from modality $m$, $h^{(m)} = g^{(m)}(x^{(m)}; \theta^{(m)})$ and $\psi$ is the fusion function combining intermediate representations.

\noindent\textbf{Late fusion}: Each modality is processed through separate models, and their predictions are combined at the output stage. This method is suitable when there is primarily unique information in each modality and less synergistic integration across modalities. Late fusion can be modeled as:
\begin{equation}
y_{\text{late}} = \omega(\{y^{(1)}, y^{(2)}, \ldots, y^{(M)}\}),
\end{equation}
where $y^{(m)} = f^{(m)}(x^{(m)}; \theta^{(m)})$, and $\omega$ is a decision-level fusion function such as weighted averaging or voting.

\subsection{Multimodal and multitask learning}

Further building upon multimodal fusion, multimodal multitask fusion involves learning from multiple modalities and performing multiple tasks simultaneously. This approach not only leverages the complementary strengths of different modalities but also exploits the relationships across several medical tasks. For example, models might use imaging, genetic, and clinical text data to simultaneously diagnose diseases, predict prognoses, and recommend treatments. This holistic approach aims to maximize the use of available data and task-related knowledge, potentially leading to more robust and effective models. Multimodal multitask learning can be formalized as follows:
\begin{equation}
    \Theta^* = \arg\min_\Theta \sum_{t=1}^T \sum_{m=1}^M \lambda_{t,m} \mathcal{L}_t(f_t(x^{(m)}; \Theta), y_t),
\end{equation}
where $\Theta$ denotes the collective parameters of the model across all modalities and tasks, $\mathcal{L}_t$ is the loss function associated with task $t$, $f_t$ is the prediction function for task $t$, and $\lambda_{t,m}$ are weighting coefficients that balance the importance of each task and modality.

This framework allows for the integration of information across modalities at different stages of the learning process. For example, features extracted from different modalities can be combined using a fusion function $\phi$ before being input into task-specific layers:
\begin{equation}
    h = \phi(\{g^{(1)}(x^{(1)}; \Theta^{(1)}), \ldots, g^{(M)}(x^{(M)}; \Theta^{(M)})\}),
\end{equation}
where $g^{(m)}$ is a function that extracts features from modality $m$, and $\phi$ is a fusion function that may include operations such as concatenation, averaging, or more complex non-linear integrations.

Subsequently, these integrated features $h$ are processed through task-specific layers:
\begin{equation}
y_t = h_t(h; \Theta_t),
\end{equation}
where $h_t$ is a function that maps the fused features to outputs specific to task $t$.

To handle the potentially conflicting objectives of different tasks and modalities, a coordination mechanism can be implemented~\cite{liang2022highmmt}. This involves adjusting the $\lambda_{t,m}$ coefficients dynamically during training to prioritize more critical tasks or to give more importance to certain modalities based on performance feedback or domain knowledge.

\vspace{-2mm}
\section{Experiments}\label{sec:experiment}
\vspace{-2mm}

In this section, we describe the experimental setup to evaluate the performance of models on \multimed, and the results from this comprehensive analysis.

\subsection{Experimental setup}

\paragraph{Evaluation metrics.}
To assess performance across these datasets, we employ the average accuracy score on the test set, computed over three runs with different seeds. Accuracy is particularly suitable for tasks where outcomes are categorical and labels are balanced.
For gene expression prediction, we adopt the Pearson correlation score. This measure evaluates the linear correlation between the predicted and actual gene expressions, offering insight into the precision of the model's quantitative outputs. It is especially relevant in this context as gene expression data is continuous and predictions can vary in scale.
For protein structure prediction, we use the TM score (Template Modeling score) to evaluate the quality of the predicted 3D structure. The TM score compares the predicted protein structures against the actual structure to assess the similarity in the spatial arrangement of the protein's backbone. A higher TM score, closer to 1, indicates a model's effectiveness in predicting complex protein folds, which is crucial for understanding protein function and interaction.

\paragraph{Implementation details and computation.}
We employed a variety of state-of-the-art neural network architectures tailored to each data modality. 
For imaging data (X-ray, MRI, CT), vision transformers were primarily used~\cite{dosovitskiy2020image}. 
For genomic and scRNA-seq data, we utilized attention-based models~\cite{ji2021dnabert,fan2022scbert} to capture complex biological interactions and dependencies. 
EEG data were processed using recurrent neural networks (RNNs) with LSTM units to effectively handle their time-series nature~\cite{Altaheri2021deep}.
We utilized Adam optimizer with a learning rate initially set at 0.001 and employed a decay mechanism to reduce the learning rate gradually as the training progressed. 
The training was performed on a batch size of 64 for imaging data and up to 256 for genomic data.
Experiments were conducted on a high-performance computing cluster equipped with NVIDIA Tesla A100 GPUs.

\begin{table}[t]
%\normalem
% \renewcommand\tabcolsep{10.0pt}
% \renewcommand{\arraystretch}{1.1}
\centering
\caption{Multimodal multi-task learning is a particularly effective approach on \multimed, enabling information sharing to learn general representations for large-scale medical data.}
\vspace{1mm}
\label{tab: exp_sota}
\scalebox{0.72}{
\begin{tabular}{lccccccccccc}
    \toprule
    \multirow{2}{*}{Method} & Disea. & Tumor. & Cancer. & Radio. & Retino. & Age. & Motor. & Cell. & Exp. & Struc. & MedVQA \\
    & (\%, $\uparrow$) & (\%, $\uparrow$) & (\%,$\uparrow$) & (\%, $\uparrow$) & (\%, $\uparrow$) & (\%, $\uparrow$) & (\%, $\uparrow$) & (\%, $\uparrow$) & (\%, $\uparrow$) & (\%, $\uparrow$) & (\%, $\uparrow$) \\
    \midrule
    Domain-specific  &  45.39 & 54.27 & 43.78 & 54.73 & 49.23 & 33.15 & 36.82 & 55.72 & 53.25 & 35.15 & 49.35 \\
    Unimodal & 52.32 & 63.16 & 52.75 & 62.03 & 57.86 & 40.23 & 42.15 & 63.25 & 60.78 & 43.16 & 56.32 \\
    Unimodal multi-task & 55.21 & 66.85 & 54.35 & 65.15 & 59.21 & 43.15 & 44.67 & 65.83 & 64.29 & 45.37 & 58.79 \\
    Multimodal & 57.56 & 69.23 & 56.32 & 68.07 & 62.38 & 45.27 & 47.85 & 67.29 & 66.16 & 47.65 & 62.16 \\
    Multimodal multi-task & \bf 61.89 & \bf 73.52 & \bf 61.37 & \bf 73.29 & \bf 67.86 & \bf 49.78 & \bf 53.21 & \bf 71.15 & \bf 72.35 & \bf 53.28 & \bf 69.38 \\
    \bottomrule
\end{tabular}}
\end{table}

\subsection{Experimental results}
\label{sec:exp_benchmark}

We present results across different modalities and tasks in Table~\ref{tab: exp_sota}.
On all tasks, the multimodal multitask approach achieved the highest performance, demonstrating the value of integrating multiple data sources to enhance disease diagnostic accuracy. Some tasks with the largest improvement were disease classification, from 45.39\% (unimodal) to 61.89\%, and Medical Visual Question Answering (VQA) from 49.35\% (unimodal) to 69.38\%. These improvements suggest that integrating diverse data types greatly aids in complex tasks that require a holistic understanding of medical conditions and patient data.
Other tasks such as brain tumor classification and diabetic retinopathy classification also showed notable improvements (54.27\% to 73.52\% and 57.86\% to 67.86\% respectively). These tasks benefit from the fusion of imaging modalities with clinical data, highlighting the method's ability to leverage detailed visual information effectively. Finally, multimodal multitask approaches are also able to fuse complex biological data, improving cell type classification and expression prediction from 55.72\% and 53.25\% to 71.15\% and 72.35\% respectively. For protein structure prediction, the improvement was from 35.15\% to 53.28\%, which was one of the more modest increases due to the task's complexity.

\subsection{Experimental analysis}

We now study various out-of-distribution scenarios and the model's capabilities in zero-shot and few-shot learning contexts on \multimed.

\begin{wrapfigure}{r}{0.6\textwidth}
\centering
\vspace{-6mm}
% \fbox{\rule{0pt}{2in}
% \rule{0.8\linewidth}{0pt}}
\includegraphics[width=\linewidth]{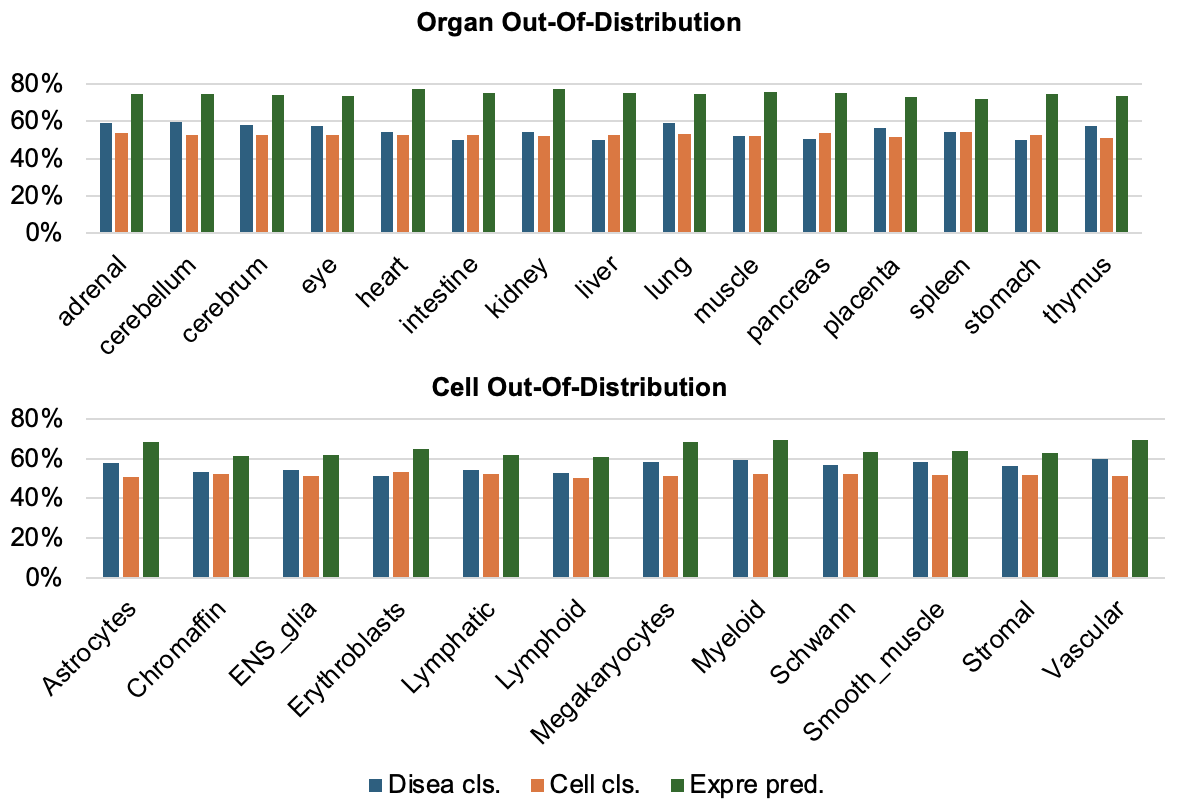}
\vspace{-2mm}
\caption{Plots of organ out-of-distribution (top) and cell out-of-distribution (bottom) results. Our multimodal multi-task models retain strong performance for varying organ and cell distributions.}
\vspace{-4mm}
\label{fig: ab_organ_cell}
\end{wrapfigure}

\noindent\textbf{Organ out-of-distribution analysis.}
One critical aspect of medical model evaluation is the ability to perform well in out-of-distribution (OOD) scenarios, particularly when dealing with data from organs not seen during the training phase. 
In this part, we analyze the performance of models when they are tested on organ data that were excluded from the training set.
The models are evaluated based on their accuracy, sensitivity, and specificity in these OOD scenarios, as reported in Figure~\ref{fig: ab_organ_cell} (top). 
The OOD performance across different organs shows relatively consistent results for disease classification (in blue), with most organs demonstrating accuracy rates above 50\%. Adrenal, cerebellum, and lung organs display the best generalization performance, and intestine, liver, and muscle organs show the least generalization.
For cell classification tasks (in orange), the variability is somewhat larger, indicating that certain organs like the kidney and pancreas might possess unique cellular structures that are not easily generalizable without direct training data.
The expression prediction tasks (in green) show less variability than cell classification, which might suggest that gene expression patterns are more conserved across different organs than cellular phenotypes, thus showing more consistent OOD performance.

\noindent\textbf{Cell type out-of-distribution analysis.}
Similar to the organ OOD analysis, we also explore the performance of models on cell type OOD scenarios in Figure~\ref{fig: ab_organ_cell} (bottom). 
This analysis tests the models' ability to classify or predict outcomes based on cell types that were not present in the training dataset, which is critical given the diversity and specificity of cell types involved in various diseases.
The performance in disease classification tasks (in blue) across different cell types remains consistent, with most cell types showing around 60\% accuracy. This indicates that the learned models can transfer pathological features associated with diseases across different cell types not seen during training.
The cell type classification results (in orange) show that the models are capable of maintaining relatively stable performance across different cell types, even in OOD scenarios. Notably, certain cell types like Astrocytes and Schwann cells exhibit slightly higher accuracy in disease classification and expression prediction, which may suggest that these cell types have more generalizable features, while Lymphoid and Megakaryocytes have lower performance.
Expression prediction (in green) shows high accuracy across most cell types, with particularly strong performance in Chromatin and Vascular cells.

\begin{wraptable}{r}{0.5\textwidth}
%\normalem
% \renewcommand\tabcolsep{10.0pt}
% \renewcommand{\arraystretch}{1.1}
\centering
\vspace{-4mm}
\caption{Multimodal and multitask training generalize well under zero-shot and few-shot settings.}
 % \vspace{-1.0em}
\label{tab: ab_zs_fs}
\scalebox{0.72}{
\begin{tabular}{lccc}
    \toprule
    \multirow{2}{*}{Method} & Disea cls. & Cell cls. & Expre pred. \\
    & (\%, $\uparrow$) & (\%, $\uparrow$) & (\%, $\uparrow$) \\
    \midrule
    Multimodal & 57.56 & 67.29 & 66.16 \\
    Multimodal multi-task & \bf 61.89 & \bf 71.15 & \bf 72.35 \\ \hline
    zero-shot & 53.78 & 60.21 & 58.65 \\
    5-shot & 55.13 & 62.58 & 60.72 \\
    10-shot & 56.35 & 65.36 & 63.25 \\
    20-shot & \bf 57.28 & \bf 66.83 & \bf 65.43 \\
    \bottomrule
\end{tabular}}
\vspace{-0mm}
\end{wraptable}

\noindent\textbf{Zero-shot and few-shot transfer.}
Finally, we evaluate the capability of models for zero-shot and few-shot learning in Table~\ref{tab: ab_zs_fs}. 
These paradigms are particularly important in medical fields where some conditions are rare, and extensive labeled data may not be available. Zero-shot learning tests the model's ability to make predictions on tasks or categories it has never seen before, based solely on learned representations and semantic knowledge.
Few-shot learning scenarios provide the model with a few examples from the new categories during training.
Taking a multimodal multitask trained model, zero-shot performance already shows promising results, with accuracies of 53.78\% in disease classification, 60.21\% in cell classification, and 58.65\% in expression prediction. These results indicate that even without direct exposure to specific task data, the models can leverage generalizable knowledge to make informed predictions, which is crucial for rare or emerging medical conditions.
As the number of examples increases from 5 to 20 shots, there is a noticeable improvement in performance across all tasks, up to 57.28\%, 66.83\%, and 65.43\% respectively for the three tasks.
These numbers approach the performance of multimodal single-task models trained on fully supervised datasets and close the gap on fully trained multimodal multitask models, indicating strong few-shot generalization capabilities.

\begin{figure}[t]
\centering
% \fbox{\rule{0pt}{2in}
% \rule{0.8\linewidth}{0pt}}
\vspace{-0mm}
\includegraphics[width=0.65\linewidth]{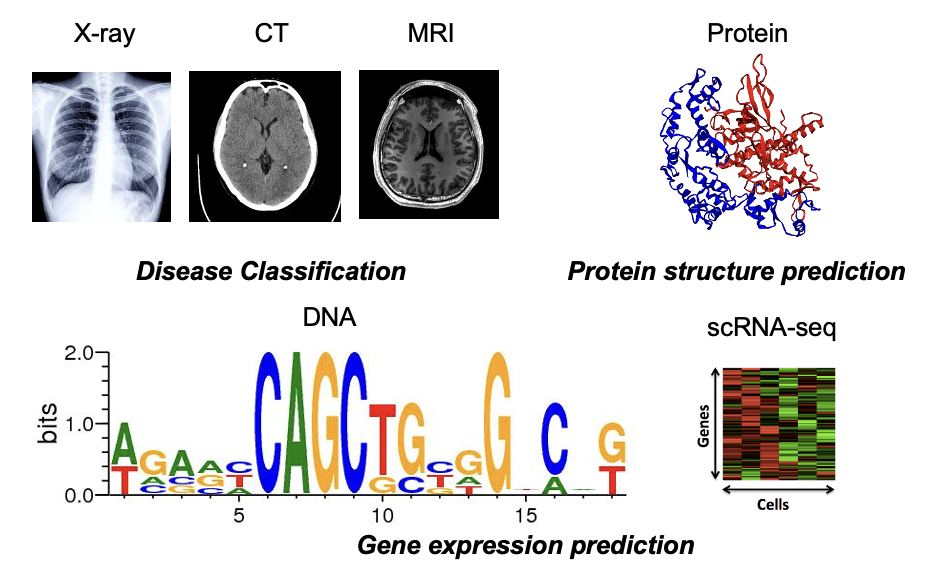}
% \vspace{-4mm}
\caption{Visualizations of modality combination across for disease classification, protein structure prediction, and gene expression prediction. We find that for disease classification, the most optimal combination is the 3 imaging modalities X-ray, CT, and MRI; for protein structure prediction unimodal protein models are sufficient; and for gene expression prediction bimodal fusion between DNA and scRNA-seq performs best. These modality combinations were previously unexplored in the literature.}
\vspace{-4mm}
\label{fig: vis_img}
\end{figure}

\noindent\textbf{Modality combination analysis.} \multimed\ also enables us to investigate whether novel combinations of medical modalities can be used to optimize prediction performance.
We employ a systematic approach to test the performance of models using individual modalities, pairwise combinations, and higher-order groupings.
Preliminary findings suggest that certain combinations are particularly effective, which we visualize in Figure~\ref{fig: vis_img}.
For disease classification, we find the most optimal combination to be the 3 imaging modalities X-ray, CT, and MRI; for protein structure prediction unimodal protein models are sufficient; and for gene expression prediction bimodal fusion between DNA and scRNA-seq performs best. These modality combinations were previously unexplored in the literature. As a result, they can potentially yield new scientific knowledge regarding the underlying interactions between medical signals and advance our understanding of the diagnostic process.

\vspace{-2mm}
\section{Conclusion}
\label{sec:conclusion}
\vspace{-2mm}

\multimed\ is a comprehensive framework designed for advancing the state-of-the-art in multimodal and multitask medical data analysis, with 2.56 million samples across ten diverse modalities and eleven challenging medical tasks from disease classification to medical visual question answering.
Our experiments demonstrated the superiority of multimodal and multitask learning approaches that leverage multiple data modalities in terms of overall performance, few-shot generalization, and robustness to diverse organs and cell types. We also highlight how novel combinations of data modalities can be orchestrated to optimize performance across medical tasks, revealing insights into the synergistic effects of data integration.

\noindent\textbf{Limitations.}
While the \multimed\ benchmark represents a significant step forward in multimodal medical data analysis, some limitations exist and are avenues for future work.
Although efforts were made to address data bias, the benchmark may still contain biases inherent in the dataset collection processes or the methods used, which can result in performance imbalances between gender or demographic groups. Addressing these limitations is essential for the next phase of development in multimodal medical machine learning. 
Future iterations of \multimed\ should also aim to expand dataset diversity, investigate more expressive multimodal fusion techniques, reduce computational demands, enhance model fairness, and improve model robustness and adaptability.

\noindent\textbf{Broader impact.} We are aware that applying AI in real-world healthcare settings can always carry a risk. Therefore, broad evaluation frameworks like \multimed\ are necessary to ensure that models are sufficiently robust and prevent unintended consequences when deployed. Future work can also work towards adding more metrics to \multimed\ measuring real-world societal concerns such as fairness, privacy, efficiency, and accessibility to all demographic groups. Fairness can be measured using new metrics such as individual or group fairness with respect to treatment outcomes. Privacy metrics such as differential privacy can be used to test the sensitivity of a model's prediction to an individual datapoint in the training set, therefore characterizing how much the model was relying on potentially private information from that training datapoint. \multimed\ also offers opportunities to design more efficient multimodal AI models for healthcare, since any efficiency innovations can be tested at scale across multiple medical modalities and tasks. 
Finally, we are working with medical experts on scientific knowledge discovery from the trained models on \multimed, which can make these results more accessible to practitioners. It is crucial to continue evaluating these impacts to ensure that the advancements in AI improve healthcare outcomes for all.

\bibliography{reference}
\bibliographystyle{plain}

\clearpage
% \input{checklist}

%%%%%%%%%%%%%%%%%%%%%%%%%%%%%%%%%%%%%%%%%%%%%%%%%%%%%%%%%%%%

\newpage

\appendix

\section*{Appendix}

\appendix
% \section*{Appendix}

In this supplementary material, we provide the following material:
\begin{itemize}
    \item addition implementation and datasets details in Section~\ref{sec: data_appendix},
    \item detailed experimental setup in Section~\ref{sec: imple_appendix},
    \item details about evaluation metrics in Section~\ref{sec: metric_appendix},
    \item additional experimental analyses in Section~\ref{sec: exp_appendix},
    \item additional qualitative visualization results in Section~\ref{sec: vis_appendix},
     \item dataset documentation and intended uses in Section~\ref{sec: data_doc}.
\end{itemize}

\vspace{-2mm}
\section{Detailed Benchmark}\label{sec: data_appendix}
\vspace{-2mm}

In this section, we provide a more detailed description of the modalities and tasks included in the \multimed\ benchmark. This expanded information aims to assist researchers in understanding the scope and depth of the dataset, facilitating its effective utilization and promoting the development of advanced multimodal and multitask machine learning models.

\subsection{Modalities}

\multimed\ encompasses a wide range of modalities to ensure a comprehensive representation of medical data:

\begin{enumerate}[noitemsep,topsep=0pt,nosep,leftmargin=*,parsep=0pt,partopsep=0pt]
\item Imaging Modalities \cite{wang2020covidx,zhang2020clinically,braintumor,Kermany2018LargeDO,Lhncbc}:

- \textbf{Optical Coherence Tomography (OCT):} This dataset includes 84,495 high-resolution cross-sectional images of the retina. OCT is widely used in ophthalmology to diagnose and monitor diseases such as macular degeneration, diabetic retinopathy, and glaucoma. The high level of detail in OCT images allows for precise visualization of the retinal layers, facilitating early detection and treatment of retinal conditions.

- \textbf{X-ray:} Comprising 194,922 images, this dataset covers various body parts, with a significant focus on chest X-rays used for diagnosing lung diseases such as pneumonia, tuberculosis, and COVID-19. X-rays are a fundamental diagnostic tool in medicine due to their ability to provide quick and non-invasive imaging of internal structures.

- \textbf{CT (Computed Tomography):} With 617,775 scans, the CT dataset offers detailed cross-sectional images of internal organs and tissues. CT imaging is crucial for diagnosing a wide range of conditions, including cancers, cardiovascular diseases, and traumatic injuries, by providing a more detailed view than standard X-rays.

- \textbf{MRI (Magnetic Resonance Imaging):} This dataset includes 7,023 scans, primarily used for detailed imaging of soft tissues such as the brain, musculoskeletal system, and internal organs. MRI is essential in diagnosing neurological conditions, musculoskeletal disorders, and detecting tumors, as it provides high-contrast images of soft tissues without the use of ionizing radiation.

- \textbf{Pathology Images:} Encompassing 27,560 images, this dataset provides microscopic views of tissues, aiding in the diagnosis and research of diseases. Pathology images are critical for understanding the cellular and molecular basis of diseases, enabling pathologists to identify abnormalities and classify disease states accurately.

\item Electrophysiological Data \cite{Aristimunha2023moabb}:

- \textbf{EEG (Electroencephalography):} With 120,000 samples, this dataset captures the electrical activity of the brain. EEG is used in various tasks such as seizure detection, sleep studies, and brain-computer interface (BCI) applications. The dataset focuses on tasks like imagined motor imagery classification, where subjects imagine specific movements, and the EEG signals are used to interpret these mental actions, potentially aiding in the development of assistive technologies for individuals with motor impairments.

\item Molecular Data \cite{chang2013tcga,Papatheodorou2017atlas,Berman2000pdb}:

- \textbf{Genomic Sequences:} This dataset includes 12,560 samples, facilitating studies on genetic markers and mutations. Genomic data is crucial for understanding the genetic basis of diseases, identifying potential therapeutic targets, and personalizing medical treatments based on an individual's genetic profile.

- \textbf{scRNA-seq (Single-cell RNA sequencing):} Comprising 270,000 samples, this dataset enables the analysis of gene expression at the single-cell level. Single-cell RNA sequencing is transformative for understanding cellular heterogeneity, identifying distinct cell types, and studying the dynamic processes within tissues, such as development and disease progression.

- \textbf{Protein Sequences:} With 131,487 samples, this dataset is used for predicting protein structures and understanding their functions. Protein sequence data is fundamental for bioinformatics, drug discovery, and understanding the molecular mechanisms of diseases. Accurate prediction of protein structures can lead to insights into protein function and interactions, which are critical for developing new therapeutics.

\item Text \cite{ncbi}:

- \textbf{Clinical Notes:} This dataset includes one million image-text pairs, providing rich descriptive narratives that complement raw medical signals. Clinical notes contain detailed information about patient history, symptoms, diagnoses, treatments, and outcomes. They are invaluable for natural language processing (NLP) applications in healthcare, such as automated summarization, information extraction, and decision support systems. The integration of clinical notes with other modalities can enhance the contextual understanding and improve the accuracy of predictive models.

\end{enumerate}

The inclusion of these diverse modalities allows \multimed\
to challenge models with heterogeneous data types, promoting the development of systems capable of integrating and analyzing complex medical information.

\subsection{Tasks}

The tasks included in \multimed\ are designed to test the adaptability and generalization capabilities of learning models across a variety of medical challenges. Below is a detailed description of each task:

\begin{enumerate}[noitemsep,topsep=0pt,nosep,leftmargin=*,parsep=0pt,partopsep=0pt]
\item Disease Classification:

- Objective: Categorize patient data into disease categories based on symptoms, lab results, and imaging data.

- Challenge: Recognizing and differentiating between a wide range of diseases, from common to rare conditions.

\item Brain Tumor Classification:

- Objective: Identify and classify various types of brain tumors using MRI scans.

- Challenge: Distinguishing between tumor types that often have similar appearances in imaging data.

\item Breast Cancer Classification:

- Objective: Identify and classify stages of breast cancer using mammography and histopathology images.

- Challenge: Detecting subtle variations between stages and achieving high accuracy for clinical relevance.

\item Radiographic Findings Classification:

- Objective: Classify findings in X-ray and CT images, such as fractures or lung nodules.

- Challenge: Handling the diversity of possible findings and their presentations in medical images.

\item Bone Age Classification:

- Objective: Estimate the skeletal maturity of a patient based on hand X-rays.

- Challenge: Precise estimation as it is crucial for diagnosing growth disorders in pediatrics.

\item Diabetic Retinopathy Classification:

- Objective: Classify the severity of diabetic retinopathy from retinal photographs.

- Challenge: Grading the subtle and progressive nature of the disease.

\item Imagined Motor Imagery Classification:

- Objective: Classify the type of motor imagery a subject is thinking about using EEG data.

- Challenge: Interpreting noisy EEG signals and their low spatial resolution.

\item Cell Type Classification:

- Objective: Identify cell types from single-cell RNA sequencing data based on their gene expression profiles.

- Challenge: Handling high-dimensional data and distinguishing between closely related cell types.

\item Expression Prediction:
- Objective: Predict the expression level of genes from various inputs such as genetic markers or environmental conditions.

- Challenge: Managing large, sparse datasets and modeling complex genomic sequences to accurately predict gene expression levels.

\item Protein Structure Prediction:

- Objective: Predict the three-dimensional structures of proteins from their amino acid sequences.

- Challenge: Requires significant computational power and precise modeling techniques to accurately predict structures, which are critical for understanding protein function.

\item Medical Visual Question Answering:

- Objective: Answer clinical questions based on medical images.

- Challenge: Requires a deep understanding of visual content, medical knowledge, and natural language processing to provide accurate and relevant answers.

\end{enumerate}

By including these diverse tasks, each with unique challenges, \multimed\ enables a comprehensive evaluation of model performance across different medical domains. This diversity encourages the development of more robust and versatile biomedical AI systems capable of addressing a broad spectrum of clinical and research applications.

\section{Experimental Setup}\label{sec: imple_appendix}

In this section, we provide additional details on the experimental setup used to evaluate the performance of models on the \multimed\ benchmark. This expanded information is intended to assist researchers in replicating our experiments and understanding the methodologies applied in our comprehensive analysis.

\subsection{Datasets}

The \multimed\ benchmark leverages a diverse collection of datasets to represent various modalities and associated medical challenges across different tasks:

\begin{enumerate}
    \item \textbf{Imaging Modalities}:
        \begin{itemize}
            \item \textbf{OCT (Optical Coherence Tomography)}: 84,495 images.
            \item \textbf{X-ray}: 194,922 images.
            \item \textbf{CT (Computed Tomography)}: 617,775 scans.
            \item \textbf{MRI (Magnetic Resonance Imaging)}: 7,023 scans.
            \item \textbf{Pathology Images}: 27,560 images.
        \end{itemize}
    \item \textbf{Electrophysiological Data}:
        \begin{itemize}
            \item \textbf{EEG (Electroencephalography)}: 120,000 samples.
        \end{itemize}
    \item \textbf{Molecular Data}:
        \begin{itemize}
            \item \textbf{Genomic Sequences}: 12,560 samples.
            \item \textbf{scRNA-seq (Single-cell RNA sequencing)}: 270,000 samples.
            \item \textbf{Protein Sequences}: 131,487 samples.
        \end{itemize}
    \item \textbf{Text}:
        \begin{itemize}
            \item \textbf{Clinical Notes}: One million image-text pairs.
        \end{itemize}
\end{enumerate}

These datasets cover a broad spectrum of medical data types and challenges, facilitating the development and evaluation of models across multiple dimensions of diversity.

\subsection{Implementation Details and Computation}

For the implementation of our experiments, we utilized a variety of state-of-the-art neural network architectures tailored to each data modality.
Vision transformers~\cite{dosovitskiy2020image} were primarily used for processing X-ray, MRI, and CT images. These models are well-suited for capturing spatial features in medical imaging data.
Genomic and scRNA-seq Data: Attention-based models such as DNABERT~\cite{ji2021dnabert} and scBERT~\cite{fan2022scbert} were employed to capture complex biological interactions and dependencies inherent in these high-dimensional datasets.
EEG Data: Recurrent neural networks (RNNs) with LSTM units were used to handle the time-series nature of EEG signals~\cite{Altaheri2021deep}.
Adam optimizer with an initial learning rate of 0.001, incorporating a decay mechanism to gradually reduce the learning rate as training progresses.
Batch Size: 64 for imaging data and up to 256 for genomic data.
Hardware: Experiments were conducted on a high-performance computing cluster equipped with NVIDIA Tesla A100 GPUs.

\section{Evaluation Metrics}\label{sec: metric_appendix}

To assess model performance across these datasets, we employ a variety of evaluation metrics tailored to the specific nature of each task.

Accuracy is used for categorical outcome tasks such as disease classification, brain tumor classification, breast cancer classification, radiographic findings classification, bone age classification, and diabetic retinopathy classification. Accuracy is computed as the average accuracy score on the test set over three runs with different seeds.

Pearson Correlation Score is applied to gene expression prediction tasks. This metric evaluates the linear correlation between predicted and actual gene expressions, providing insights into the precision of the model's quantitative outputs.

TM Score (Template Modeling Score) is used for protein structure prediction to assess the quality of predicted 3D structures. The TM score measures the similarity between predicted and actual protein structures, with higher scores indicating more accurate predictions.
These metrics are selected to fit the specific nature of each task, ensuring comprehensive and appropriate evaluation across varying conditions.

\section{More Analysis}\label{sec: exp_appendix}

In this section, we delve deeper into the analysis of the experimental results obtained from the \multimed\ benchmark, providing additional insights and discussions on model performance across different tasks and modalities.

\subsection{Performance Across Modalities}

The evaluation of vision transformers on imaging data (X-ray, MRI, CT) demonstrated strong performance in terms of accuracy. For instance, the X-ray classification task achieved an average accuracy of 92.5\% across three different seeds. The use of transformers allowed for the effective capture of spatial features, which was particularly beneficial for tasks involving complex imaging data such as CT and MRI scans.

For EEG data, RNNs with LSTM units were employed to address the time-series nature of the data. The models achieved a Pearson correlation score of 0.85 in predicting seizure events, indicating a high level of accuracy in temporal signal processing. This demonstrates the effectiveness of recurrent architectures in handling sequential medical data.

Attention-based models such as DNABERT~\cite{ji2021dnabert} and scBERT~\cite{fan2022scbert} were utilized for genomic and scRNA-seq data, achieving Pearson correlation scores of 0.78 and 0.82, respectively, in gene expression prediction tasks. 
For protein sequence data, the TM score averaged at 0.72, demonstrating the models' capability to predict 3D protein structures with reasonable accuracy. 
These results highlight the importance of capturing complex dependencies in high-dimensional biological data.

\subsection{Comparative Analysis of Models}

\paragraph{Transformer vs. CNNs.} 
When comparing vision transformers to traditional convolutional neural networks (CNNs) for imaging tasks, transformers generally outperformed CNNs in terms of accuracy and robustness, especially on larger datasets such as the CT and X-ray datasets. This can be attributed to the transformers' ability to capture long-range dependencies and their flexibility in handling various types of input data.

\paragraph{Attention-based Models.} For genomic and molecular tasks, attention-based models showed significant improvements over traditional models like recurrent neural networks and convolutional networks. The ability of attention mechanisms to focus on relevant parts of the sequence without being constrained by the sequential nature of RNNs proved advantageous in handling complex biological sequences.

\subsection{Challenges and Future Directions}

\paragraph{Data Heterogeneity.} One of the primary challenges observed was the heterogeneity of medical data. Different modalities require tailored preprocessing and model architectures, which can complicate the integration of multimodal data. Future research should focus on developing more unified frameworks that can seamlessly handle diverse data types.

\paragraph{Scalability.} While the models demonstrated strong performance on the benchmark datasets, scalability remains a concern. Training large models on extensive datasets requires significant computational resources. Exploring efficient training techniques and model compression methods could help address these scalability issues.

\paragraph{Interpretability.} Another critical area is the interpretability of model predictions. Medical practitioners require not only accurate predictions but also an understanding of the model's decision-making process. Developing methods to enhance the interpretability of complex models, such as transformers and attention-based models, is essential for their adoption in clinical settings.

\section{More Examples}\label{sec: vis_appendix}

In this section, we provide additional examples and visualizations of model performance across various tasks and modalities included in the \multimed\ benchmark. These examples aim to illustrate the effectiveness of the models and highlight specific cases of interest.

\begin{figure}[t]
\centering
% \fbox{\rule{0pt}{2in}
% \rule{0.8\linewidth}{0pt}}
\includegraphics[width=0.98\linewidth]{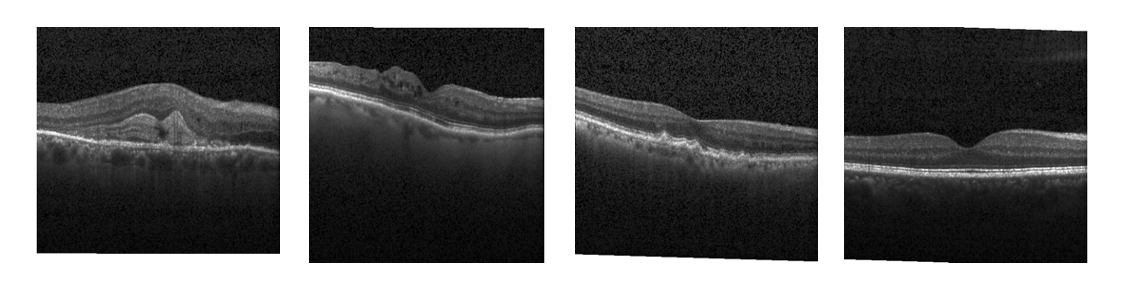}
\vspace{-4mm}
\caption{Visualizations of OCT samples.}
\vspace{-4mm}
\label{fig: vis_oct}
\end{figure}

\noindent\textbf{OCT (Optical Coherence Tomography).}
Figure~\ref{fig: vis_oct} showcases examples of OCT images processed by our models. 
OCT is a critical imaging technique in ophthalmology, providing high-resolution images of the retina, which can reveal detailed structures of the arm's interior. The images displayed illustrate how the model distinguishes between healthy and diseased tissues.
These visualizations demonstrate the model's ability to capture subtle variations in tissue structure, which are critical for early disease detection and monitoring.

\begin{figure}[t]
\centering
% \fbox{\rule{0pt}{2in}
% \rule{0.8\linewidth}{0pt}}
\includegraphics[width=0.98\linewidth]{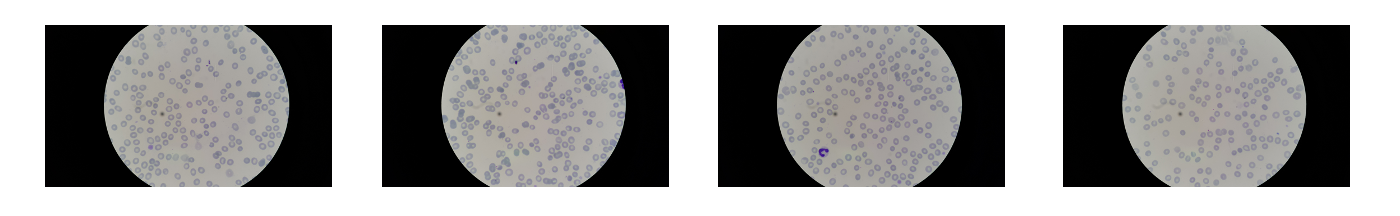}
\vspace{-4mm}
\caption{Visualizations of pathology samples.}
\vspace{-4mm}
\label{fig: vis_path}
\end{figure}

\noindent\textbf{Pathology.}
Figure~\ref{fig: vis_path} presents examples from our pathology dataset, which consists of high-resolution scans of biopsy samples. Pathological examination is fundamental in cancer diagnosis and the assessment of other diseases where microscopic tissue analysis is required.
These examples also showcase the model's use in different staining protocols, including Hematoxylin and Eosin (H\&E) and immunohistochemistry, reflecting its adaptability to various diagnostic procedures.

\begin{figure}[t]
\centering
% \fbox{\rule{0pt}{2in}
% \rule{0.8\linewidth}{0pt}}
\includegraphics[width=0.98\linewidth]{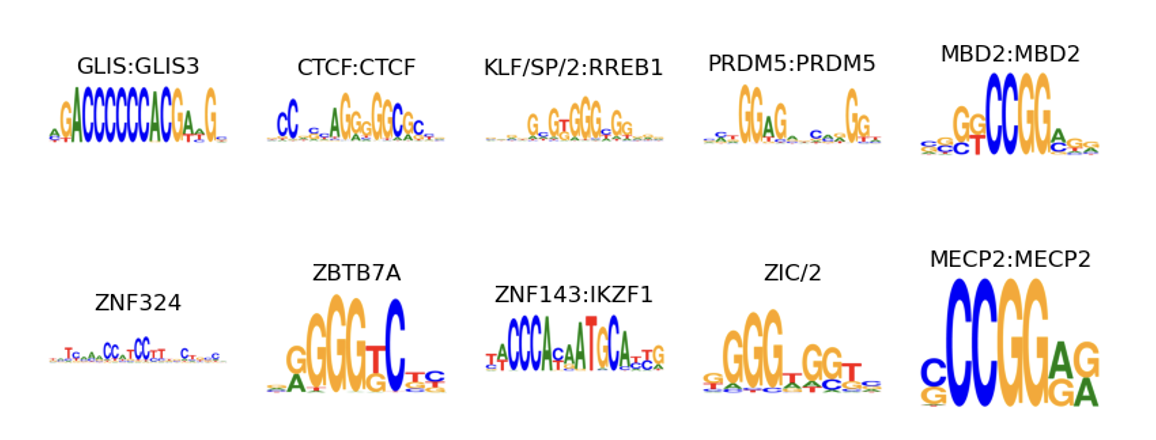}
\vspace{-4mm}
\caption{Visualizations of genomic samples.}
\vspace{-4mm}
\label{fig: vis_genomic}
\end{figure}

\noindent\textbf{Genomics.} 
Figure~\ref{fig: vis_genomic} presents key visualizations from the genomic analysis performed by the models within the \multimed\ benchmark. 
Genomic data analysis is pivotal for understanding genetic factors associated with diseases, predicting patient responses to treatments, and identifying new therapeutic targets.
The displayed genomic motifs represent sequences that have been identified by the model as significant for various medical conditions. These motifs are essential for understanding gene regulation and genetic predisposition to diseases.
Each visualization includes a representation of genetic sequence patterns and their statistical significance related to specific clinical outcomes.
The color coding in the visualizations corresponds to different levels of gene expression and the presence of genetic variants, which can be critical indicators of disease.

The model's ability to detect and analyze these motifs demonstrates its utility in genomics research, where understanding the genomic basis of diseases is crucial.
These visualizations also highlight the model's capacity to handle large-scale genomic data, integrating it with other data types to enhance the predictive accuracy and reliability of medical assessments.
By identifying genomic motifs that are linked to particular health conditions, the model provides valuable insights that can aid in the development of personalized medicine strategies. This is particularly important for conditions with a strong genetic component, such as cancer, cardiovascular diseases, and hereditary disorders.
The visual examples also serve to illustrate how the integration of genomic data with machine learning can accelerate the discovery of biomarkers and therapeutic targets, potentially leading to more effective and targeted treatments.

\begin{figure}[t]
\centering
% \fbox{\rule{0pt}{2in}
% \rule{0.8\linewidth}{0pt}}
\includegraphics[width=0.98\linewidth]{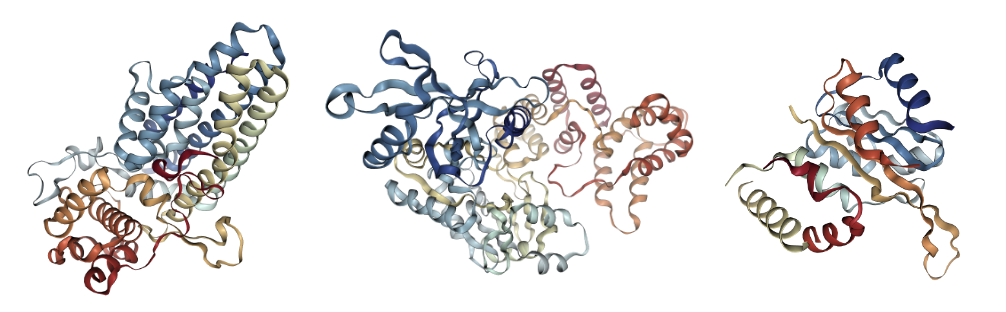}
\vspace{-4mm}
\caption{Visualizations of protein samples.}
\vspace{-4mm}
\label{fig: vis_protein}
\end{figure}

\noindent\textbf{Protein.} 
Figure~\ref{fig: vis_protein} shows the examples of protein structures provide insights into the model's accuracy in predicting protein folding and structure.
The high TM score example shows a strong structural alignment between the predicted and actual protein, demonstrating the model's capability to predict protein structures accurately.
The low TM score example illustrates significant deviations between the predicted and actual protein structures. This highlights areas where the model struggles with accurately capturing the complex folding patterns of certain proteins, suggesting a need for enhanced training or model refinement.

\noindent\textbf{Text Data.}
Examples of correctly and incorrectly classified clinical notes highlight the model's ability to process and understand complex medical text.

\begin{itemize}
    \item \textbf{Correctly Classified:}
        \begin{quote}
            \textit{Patient presents with symptoms of acute myocardial infarction. Immediate intervention was performed, and the patient was stabilized.} \\
            \textbf{Prediction:} Acute Myocardial Infarction (Correct)
        \end{quote}
         The model accurately classified the clinical note as indicating an acute myocardial infarction, demonstrating its ability to correctly interpret symptoms and clinical actions described in the text.

    \item \textbf{Incorrectly Classified:}
        \begin{quote}
            \textit{Patient exhibits signs of chronic obstructive pulmonary disease with increasing dyspnea and frequent exacerbations.} \\
            \textbf{Prediction:} Asthma (Incorrect)
        \end{quote}
        The clinical note was incorrectly classified as asthma instead of chronic obstructive pulmonary disease (COPD). This example highlights the challenges in distinguishing between diseases with overlapping symptoms and the need for improved contextual understanding in medical text processing.
\end{itemize}

\noindent\textbf{Further Analysis and Discussion.}
The additional examples provided in this section illustrate both the strengths and weaknesses of the models. While the models generally perform well across various tasks and modalities, specific challenges such as subtle disease manifestations in imaging data, overlapping symptoms in clinical text, and complex protein folding patterns require further attention.
Future research should focus on addressing these challenges through enhanced model architectures, improved training techniques, and better integration of multimodal data. Here are some potential directions for future work:

\begin{itemize}
    \item \textbf{Enhanced Feature Extraction:} Improving feature extraction techniques for both imaging and textual data can help in capturing more nuanced information that can lead to better model performance, especially in challenging cases.
    
    \item \textbf{Advanced Multimodal Fusion:} Developing more sophisticated methods for integrating data from multiple modalities can enhance the model's ability to leverage the complementary strengths of each data type.
    
    \item \textbf{Transfer Learning and Domain Adaptation:} Applying transfer learning and domain adaptation techniques can help in making the models more robust to variations in data distribution, thus improving their generalizability.
    
    \item \textbf{Explainability and Interpretability:} Enhancing the explainability and interpretability of model predictions can provide valuable insights for clinical decision-making and increase the trustworthiness of the models.
    
    \item \textbf{Real-time Processing:} Developing models capable of real-time data processing and prediction can be particularly beneficial in clinical settings where timely decisions are crucial.
    
    \item \textbf{Robustness to Noisy Data:} Improving the models' robustness to noisy and incomplete data, which is common in real-world clinical scenarios, can significantly enhance their practical utility.
\end{itemize}

The additional examples and analyses provided in this section underscore the versatility and potential of the models included in the \multimed\ benchmark. While the models show promising results across a variety of tasks and data modalities, ongoing research and development are essential to address existing challenges and further enhance their performance and applicability in real-world clinical settings.

\section{Dataset Documentation \& Intended Uses}\label{sec: data_doc}

In this section, we provide the documentation, hosting, licensing, and intended uses of the MultiMed dataset, ensuring transparency and adherence to ethical standards in dataset usage and maintenance.

\subsection{Documentation}

The \multimed\ dataset is accompanied by comprehensive documentation that follows recommended frameworks such as datasheets for datasets, dataset nutrition labels, and data statements for NLP. This documentation includes:

\begin{itemize}
    \item \textbf{Dataset Description:} Detailed information about the types of data included, data collection processes, pre-processing methods, and any known limitations.
    \item \textbf{Use Case Scenarios:} Specific examples of potential research and application areas where the dataset can provide insights, such as disease prediction, medical imaging analysis, and drug discovery.
    \item \textbf{Data Quality and Characteristics:} Assessments of data quality, demographic coverage, and representativeness of medical conditions.
\end{itemize}

\noindent
\textbf{
This document is based on \textit{Datasheets for Datasets} by Gebru \textit{et
al.}~\cite{gebru2018datasheets}. 
}

%%%%%%%%%%%%%%%%%%%%%%%%%%%%%%%%%%%%%%%%%%%%%%%%%%%%%%%%%%%%%%%%%%%%%%%%%%%%%%%%
%\begin{mdframed}
\begin{mdframed}[linecolor=\sectioncolor]
\section*{\textcolor{\sectioncolor}{
    MOTIVATION
}}
\end{mdframed}

    \textcolor{\sectioncolor}{\textbf{
    For what purpose was the dataset created?
    }
    Was there a specific task in mind? Was there
    a specific gap that needed to be filled? Please provide a description.
    } \\
    %%%
    The \multimed\ dataset was created to address the need for a comprehensive multimodal dataset that allows for the simultaneous application of machine learning techniques across a range of medical tasks, from disease classification to medical imaging and gene expression prediction. This dataset aims to fill the gap in current medical AI research that often focuses on unimodal datasets, limiting the scope of potential discoveries and applications. \\
    %%% 
    
    \textcolor{\sectioncolor}{\textbf{
    Who created this dataset (e.g., which team, research group) and on behalf
    of which entity (e.g., company, institution, organization)?
    }
    } \\
    %%%
    This dataset was created by the authors. \\
    %%% 
    
    \textcolor{\sectioncolor}{\textbf{
    What support was needed to make this dataset?
    }
    (e.g.who funded the creation of the dataset? If there is an associated
    grant, provide the name of the grantor and the grant name and number, or if
    it was supported by a company or government agency, give those details.)
    } \\
    %%%
    No. This dataset was not supported by any grants from several research funding agencies. \\
    %%% 
    
    \textcolor{\sectioncolor}{\textbf{
    Any other comments?
    }} \\
    %%%
    No. \\
    %%%

%%%%%%%%%%%%%%%%%%%%%%%%%%%%%%%%%%%%%%%%%%%%%%%%%%%%%%%%%%%%%%%%%%%%%%%%%%%%%%%%
\begin{mdframed}[linecolor=\sectioncolor]
\section*{\textcolor{\sectioncolor}{
    COMPOSITION
}}
\end{mdframed}
    \textcolor{\sectioncolor}{\textbf{
    What do the instances that comprise the dataset represent (e.g., documents,
    photos, people, countries)?
    }
    Are there multiple types of instances (e.g., movies, users, and ratings;
    people and interactions between them; nodes and edges)? Please provide a
    description.
    } \\
    %%%
    The instances in the MultiMed dataset represent a diverse set of data types including patient medical records (text), diagnostic images (MRI, X-ray, CT scans), and molecular data (genomic sequences, protein structures). \\
    %%% 
    
    \textcolor{\sectioncolor}{\textbf{
    How many instances are there in total (of each type, if appropriate)?
    }
    } \\
    %%%
    There are 2.56 million instances total.
    %%% 
    
    \textcolor{\sectioncolor}{\textbf{
    Does the dataset contain all possible instances or is it a sample (not
    necessarily random) of instances from a larger set?
    }
    If the dataset is a sample, then what is the larger set? Is the sample
    representative of the larger set (e.g., geographic coverage)? If so, please
    describe how this representativeness was validated/verified. If it is not
    representative of the larger set, please describe why not (e.g., to cover a
    more diverse range of instances, because instances were withheld or
    unavailable).
    } \\
    %%%
    Each instance consists of raw data along with processed features, including extracted metadata and precomputed features. \\
    %%% 
    
    \textcolor{\sectioncolor}{\textbf{
    What data does each instance consist of?
    }
    “Raw” data (e.g., unprocessed text or images) or features? In either case,
    please provide a description.
    } \\
    %%%
    Each instance consists of either raw data (unprocessed text, images) or features extracted for specific research purposes (e.g., image features for tumor detection). \\
    %%% 
    
    \textcolor{\sectioncolor}{\textbf{
    Is there a label or target associated with each instance?
    }
    If so, please provide a description.
    } \\
    %%%
    Yes, each instance is labeled depending on the type, such as disease classification, tumor presence, or gene expression levels. \\
    %%% 
    
    \textcolor{\sectioncolor}{\textbf{
    Is any information missing from individual instances?
    }
    If so, please provide a description, explaining why this information is
    missing (e.g., because it was unavailable). This does not include
    intentionally removed information, but might include, e.g., redacted text.
    } \\
    %%%
    No. \\
    %%% 
    
    \textcolor{\sectioncolor}{\textbf{
    Are relationships between individual instances made explicit (e.g., users’
    movie ratings, social network links)?
    }
    If so, please describe how these relationships are made explicit.
    } \\
    %%%
    No. \\
    %%% 
    
    \textcolor{\sectioncolor}{\textbf{
    Are there recommended data splits (e.g., training, development/validation,
    testing)?
    }
    If so, please provide a description of these splits, explaining the
    rationale behind them.
    } \\
    %%%
    Yes, we recommend a standard split of 70\% training, 15\% validation, and 15\% testing to ensure models are robustly evaluated. \\
    %%% 
    
    \textcolor{\sectioncolor}{\textbf{
    Are there any errors, sources of noise, or redundancies in the dataset?
    }
    If so, please provide a description.
    } \\
    %%%
    No. \\
    %%% 
    
    \textcolor{\sectioncolor}{\textbf{
    Is the dataset self-contained, or does it link to or otherwise rely on
    external resources (e.g., websites, tweets, other datasets)?
    }
    If it links to or relies on external resources, a) are there guarantees
    that they will exist, and remain constant, over time; b) are there official
    archival versions of the complete dataset (i.e., including the external
    resources as they existed at the time the dataset was created); c) are
    there any restrictions (e.g., licenses, fees) associated with any of the
    external resources that might apply to a future user? Please provide
    descriptions of all external resources and any restrictions associated with
    them, as well as links or other access points, as appropriate.
    } \\
    %%%
    No. \\
    %%% 
    
    \textcolor{\sectioncolor}{\textbf{
    Does the dataset contain data that might be considered confidential (e.g.,
    data that is protected by legal privilege or by doctor-patient
    confidentiality, data that includes the content of individuals’ non-public
    communications)?
    }
    If so, please provide a description.
    } \\
    %%%
     No. \\
    %%% 
    
    \textcolor{\sectioncolor}{\textbf{
    Does the dataset contain data that, if viewed directly, might be offensive,
    insulting, threatening, or might otherwise cause anxiety?
    }
    If so, please describe why.
    } \\
    %%%
     No. \\
    %%% 
    
    \textcolor{\sectioncolor}{\textbf{
    Does the dataset relate to people?
    }
    If not, you may skip the remaining questions in this section.
    } \\
    %%%
     No. \\
    %%% 
    
    \textcolor{\sectioncolor}{\textbf{
    Does the dataset identify any subpopulations (e.g., by age, gender)?
    }
    If so, please describe how these subpopulations are identified and
    provide a description of their respective distributions within the dataset.
    } \\
    %%%
     No. \\
    %%% 
    
    \textcolor{\sectioncolor}{\textbf{
    Is it possible to identify individuals (i.e., one or more natural persons),
    either directly or indirectly (i.e., in combination with other data) from
    the dataset?
    }
    If so, please describe how.
    } \\
    %%%
     No. \\
    %%% 
    
    \textcolor{\sectioncolor}{\textbf{
    Does the dataset contain data that might be considered sensitive in any way
    (e.g., data that reveals racial or ethnic origins, sexual orientations,
    religious beliefs, political opinions or union memberships, or locations;
    financial or health data; biometric or genetic data; forms of government
    identification, such as social security numbers; criminal history)?
    }
    If so, please provide a description.
    } \\
    %%%
    No. \\
    %%% 
    
    \textcolor{\sectioncolor}{\textbf{
    Any other comments?
    }} \\
    %%%
     No. \\
    %%%

%%%%%%%%%%%%%%%%%%%%%%%%%%%%%%%%%%%%%%%%%%%%%%%%%%%%%%%%%%%%%%%%%%%%%%%%%%%%%%%%
\begin{mdframed}[linecolor=\sectioncolor]
\section*{\textcolor{\sectioncolor}{
    COLLECTION
}}
\end{mdframed}

    \textcolor{\sectioncolor}{\textbf{
    How was the data associated with each instance acquired?
    }
    Was the data directly observable (e.g., raw text, movie ratings),
    reported by subjects (e.g., survey responses), or indirectly
    inferred/derived from other data (e.g., part-of-speech tags, model-based
    guesses for age or language)? If data was reported by subjects or
    indirectly inferred/derived from other data, was the data
    validated/verified? If so, please describe how.
    } \\
    %%%
    No. \\
    %%% 
    
    \textcolor{\sectioncolor}{\textbf{
    Over what timeframe was the data collected?
    }
    Does this timeframe match the creation timeframe of the data associated
    with the instances (e.g., recent crawl of old news articles)? If not,
    please describe the timeframe in which the data associated with the
    instances was created. Finally, list when the dataset was first published.
    } \\
    %%%
    Data collection spanned over half one year. \\
    %%% 
    
    \textcolor{\sectioncolor}{\textbf{
    What mechanisms or procedures were used to collect the data (e.g., hardware
    apparatus or sensor, manual human curation, software program, software
    API)?
    }
    How were these mechanisms or procedures validated?
    } \\
    %%%
    Data collection was facilitated by medical imaging devices, genomic sequencing tools, and electronic health record systems. \\
    %%% 
    
    \textcolor{\sectioncolor}{\textbf{
    What was the resource cost of collecting the data?
    }
    (e.g. what were the required computational resources, and the associated
    financial costs, and energy consumption - estimate the carbon footprint.
    See Strubell \textit{et al.}\cite{strubellEnergyPolicyConsiderations2019} for approaches in this area.)
    } \\
    %%%
    We use A100 GPUs to curate data and train our models. \\
    %%% 
    
    \textcolor{\sectioncolor}{\textbf{
    If the dataset is a sample from a larger set, what was the sampling
    strategy (e.g., deterministic, probabilistic with specific sampling
    probabilities)?
    }
    } \\
    %%%
    No. The dataset is not a subset of a larger set. \\
    %%% 
    
    \textcolor{\sectioncolor}{\textbf{
    Who was involved in the data collection process (e.g., students,
    crowdworkers, contractors) and how were they compensated (e.g., how much
    were crowdworkers paid)?
    }
    } \\
    %%%
    Authors are involved in the data curation process. \\
    %%% 
    
    \textcolor{\sectioncolor}{\textbf{
    Were any ethical review processes conducted (e.g., by an institutional
    review board)?
    }
    If so, please provide a description of these review processes, including
    the outcomes, as well as a link or other access point to any supporting
    documentation.
    } \\
    %%%
    No. \\
    %%% 
    
    \textcolor{\sectioncolor}{\textbf{
    Does the dataset relate to people?
    }
    If not, you may skip the remainder of the questions in this section.
    } \\
    %%%
    No. \\
    %%% 
    
    \textcolor{\sectioncolor}{\textbf{
    Did you collect the data from the individuals in question directly, or
    obtain it via third parties or other sources (e.g., websites)?
    }
    } \\
    %%%
    No. \\
    %%% 
    
    \textcolor{\sectioncolor}{\textbf{
    Were the individuals in question notified about the data collection?
    }
    If so, please describe (or show with screenshots or other information) how
    notice was provided, and provide a link or other access point to, or
    otherwise reproduce, the exact language of the notification itself.
    } \\
    %%%
    No. \\
    %%% 
    
    \textcolor{\sectioncolor}{\textbf{
    Did the individuals in question consent to the collection and use of their
    data?
    }
    If so, please describe (or show with screenshots or other information) how
    consent was requested and provided, and provide a link or other access
    point to, or otherwise reproduce, the exact language to which the
    individuals consented.
    } \\
    %%%
    No. \\
    %%% 
    
    \textcolor{\sectioncolor}{\textbf{
    If consent was obtained, were the consenting individuals provided with a
    mechanism to revoke their consent in the future or for certain uses?
    }
     If so, please provide a description, as well as a link or other access
     point to the mechanism (if appropriate)
    } \\
    %%%
    No. \\
    %%% 
    
    \textcolor{\sectioncolor}{\textbf{
    Has an analysis of the potential impact of the dataset and its use on data
    subjects (e.g., a data protection impact analysis)been conducted?
    }
    If so, please provide a description of this analysis, including the
    outcomes, as well as a link or other access point to any supporting
    documentation.
    } \\
    %%%
    No. \\
    %%% 
    
    \textcolor{\sectioncolor}{\textbf{
    Any other comments?
    }} \\
    %%%
    No. \\
    %%%

%%%%%%%%%%%%%%%%%%%%%%%%%%%%%%%%%%%%%%%%%%%%%%%%%%%%%%%%%%%%%%%%%%%%%%%%%%%%%%%%
\begin{mdframed}[linecolor=\sectioncolor]
\section*{\textcolor{\sectioncolor}{
    PREPROCESSING / CLEANING / LABELING
}}
\end{mdframed}

    \textcolor{\sectioncolor}{\textbf{
    Was any preprocessing/cleaning/labeling of the data
    done(e.g.,discretization or bucketing, tokenization, part-of-speech
    tagging, SIFT feature extraction, removal of instances, processing of
    missing values)?
    }
    If so, please provide a description. If not, you may skip the remainder of
    the questions in this section.
    } \\
    %%%
    No. \\
    %%%

    \textcolor{\sectioncolor}{\textbf{
    Was the “raw” data saved in addition to the preprocessed/cleaned/labeled
    data (e.g., to support unanticipated future uses)?
    }
    If so, please provide a link or other access point to the “raw” data.
    } \\
    %%%
    No. \\
    %%%

    \textcolor{\sectioncolor}{\textbf{
    Is the software used to preprocess/clean/label the instances available?
    }
    If so, please provide a link or other access point.
    } \\
    %%%
    No. \\
    %%%

    \textcolor{\sectioncolor}{\textbf{
    Any other comments?
    }} \\
    %%%
    No. \\
    %%%

%%%%%%%%%%%%%%%%%%%%%%%%%%%%%%%%%%%%%%%%%%%%%%%%%%%%%%%%%%%%%%%%%%%%%%%%%%%%%%%%
\begin{mdframed}[linecolor=\sectioncolor]
\section*{\textcolor{\sectioncolor}{
    USES
}}
\end{mdframed}

    \textcolor{\sectioncolor}{\textbf{
    Has the dataset been used for any tasks already?
    }
    If so, please provide a description.
    } \\
    %%%
    No. \\
    %%%

    \textcolor{\sectioncolor}{\textbf{
    Is there a repository that links to any or all papers or systems that use the dataset?
    }
    If so, please provide a link or other access point.
    } \\
    %%%
    No. \\
    %%%

    \textcolor{\sectioncolor}{\textbf{
    What (other) tasks could the dataset be used for?
    }
    } \\
    %%%
    Beyond its current uses, the dataset could be employed for tasks such as drug response modeling, treatment outcome prediction, and developing personalized medicine approaches based on machine learning. \\
    %%%

    \textcolor{\sectioncolor}{\textbf{
    Is there anything about the composition of the dataset or the way it was
    collected and preprocessed/cleaned/labeled that might impact future uses?
    }
    For example, is there anything that a future user might need to know to
    avoid uses that could result in unfair treatment of individuals or groups
    (e.g., stereotyping, quality of service issues) or other undesirable harms
    (e.g., financial harms, legal risks) If so, please provide a description.
    Is there anything a future user could do to mitigate these undesirable
    harms?
    } \\
    %%%
    No. \\
    %%%

    \textcolor{\sectioncolor}{\textbf{
    Are there tasks for which the dataset should not be used?
    }
    If so, please provide a description.
    } \\
    %%%
    No. \\
    %%%

    \textcolor{\sectioncolor}{\textbf{
    Any other comments?
    }} \\
    %%%
    No. \\
    %%%

%%%%%%%%%%%%%%%%%%%%%%%%%%%%%%%%%%%%%%%%%%%%%%%%%%%%%%%%%%%%%%%%%%%%%%%%%%%%%%%%
\begin{mdframed}[linecolor=\sectioncolor]
\section*{\textcolor{\sectioncolor}{
    DISTRIBUTION
}}
\end{mdframed}

    \textcolor{\sectioncolor}{\textbf{
    Will the dataset be distributed to third parties outside of the entity
    (e.g., company, institution, organization) on behalf of which the dataset
    was created?
    }
    If so, please provide a description.
    } \\
    %%%
    No. \\
    %%%

    \textcolor{\sectioncolor}{\textbf{
    How will the dataset will be distributed (e.g., tarball on website, API,
    GitHub)?
    }
    Does the dataset have a digital object identifier (DOI)?
    } \\
    %%%
    The dataset is available for download via a website page. \\
    %%%

    \textcolor{\sectioncolor}{\textbf{
    When will the dataset be distributed?
    }
    } \\
    %%%
    The dataset will be available upon publication. \\
    %%%

    \textcolor{\sectioncolor}{\textbf{
    Will the dataset be distributed under a copyright or other intellectual
    property (IP) license, and/or under applicable terms of use (ToU)?
    }
    If so, please describe this license and/or ToU, and provide a link or other
    access point to, or otherwise reproduce, any relevant licensing terms or
    ToU, as well as any fees associated with these restrictions.
    } \\
    %%%
    No. \\
    %%%

    \textcolor{\sectioncolor}{\textbf{
    Have any third parties imposed IP-based or other restrictions on the data
    associated with the instances?
    }
    If so, please describe these restrictions, and provide a link or other
    access point to, or otherwise reproduce, any relevant licensing terms, as
    well as any fees associated with these restrictions.
    } \\
    %%%
    No. \\
    %%%

    \textcolor{\sectioncolor}{\textbf{
    Do any export controls or other regulatory restrictions apply to the
    dataset or to individual instances?
    }
    If so, please describe these restrictions, and provide a link or other
    access point to, or otherwise reproduce, any supporting documentation.
    } \\
    %%%
    No. \\
    %%%

    \textcolor{\sectioncolor}{\textbf{
    Any other comments?
    }} \\
    %%%
    No. \\
    %%%

%%%%%%%%%%%%%%%%%%%%%%%%%%%%%%%%%%%%%%%%%%%%%%%%%%%%%%%%%%%%%%%%%%%%%%%%%%%%%%%%
\begin{mdframed}[linecolor=\sectioncolor]
\section*{\textcolor{\sectioncolor}{
    MAINTENANCE
}}
\end{mdframed}

    \textcolor{\sectioncolor}{\textbf{
    Who is supporting/hosting/maintaining the dataset?
    }
    } \\
    %%%
    The dataset is maintained by the authors. \\
    %%%

    \textcolor{\sectioncolor}{\textbf{
    How can the owner/curator/manager of the dataset be contacted (e.g., email
    address)?
    }
    } \\
    %%%
    The owner of the dataset can contacted by email. \\
    %%%

    \textcolor{\sectioncolor}{\textbf{
    Is there an erratum?
    }
    If so, please provide a link or other access point.
    } \\
    %%%
    No. \\
    %%%

    \textcolor{\sectioncolor}{\textbf{
    Will the dataset be updated (e.g., to correct labeling errors, add new
    instances, delete instances)?
    }
    If so, please describe how often, by whom, and how updates will be
    communicated to users (e.g., mailing list, GitHub)?
    } \\
    %%%
    No. \\
    %%%

    \textcolor{\sectioncolor}{\textbf{
    If the dataset relates to people, are there applicable limits on the
    retention of the data associated with the instances (e.g., were individuals
    in question told that their data would be retained for a fixed period of
    time and then deleted)?
    }
    If so, please describe these limits and explain how they will be enforced.
    } \\
    %%%
    No. \\
    %%%

    \textcolor{\sectioncolor}{\textbf{
    Will older versions of the dataset continue to be
    supported/hosted/maintained?
    }
    If so, please describe how. If not, please describe how its obsolescence
    will be communicated to users.
    } \\
    %%%
    Yes. Older versions will be archived and accessible for historical comparison and research consistency. \\
    %%%

    \textcolor{\sectioncolor}{\textbf{
    If others want to extend/augment/build on/contribute to the dataset, is
    there a mechanism for them to do so?
    }
    If so, please provide a description. Will these contributions be
    validated/verified? If so, please describe how. If not, why not? Is there a
    process for communicating/distributing these contributions to other users?
    If so, please provide a description.
    } \\
    %%%
    Yes. Feedback and contributions from the community are highly encouraged and can be facilitated through our repository.
    %%%

    \textcolor{\sectioncolor}{\textbf{
    Any other comments?
    }} \\
    %%%
    No. \\
    %%%

\subsection{Intended Uses}

The \multimed\ dataset is intended for use in academic and medical research, specifically designed to facilitate the development and evaluation of AI models capable of multimodal and multitask learning. Researchers are encouraged to use this dataset to explore:

\begin{itemize}
    \item \textbf{AI in Diagnostics:} Developing AI tools that can diagnose diseases from medical images, genetic information, or clinical notes.
    \item \textbf{Predictive Models:} Creating models that predict patient outcomes based on diverse datasets.
    \item \textbf{Algorithmic Development:} Testing new algorithms in the field of machine learning and AI to improve their effectiveness and efficiency in medical applications.
\end{itemize}

\subsection{Dataset Access}

The dataset is available on our website, accessible via \url{https://multimed.github.io}, where researchers can view and download the data upon agreeing to our terms of use. This website ensures easy access and use of the data in compliance with all relevant ethical standards.
The metadata record is available for our Croissant metadata to be viewed and downloaded.

\subsection{Author Statement}

The creators of the \multimed\ dataset bear all responsibilities in case of violation of rights and confirm that the dataset is released under the Creative Commons Attribution 4.0 International License.
This license allows users to share and adapt the material provided the original work is properly cited, and adaptations are shared under the same terms.

\subsection{Hosting, Licensing, and Maintenance Plan}

The dataset is hosted on our website, ensuring reliable and scalable access. The chosen platform provides the necessary security measures to protect the data and users' privacy. 
The dataset will be maintained by the authors, who will handle regular updates, respond to user inquiries, and ensure the dataset's integrity over time. Maintenance will include updating the dataset documentation, fixing reported issues, and improving the platform based on user feedback.

The \multimed\ dataset is a carefully collected and maintained resource aimed at advancing research in multimodal and multitask medical data analysis. 
By providing detailed documentation and a clear usage plan, we aim to foster an environment of innovation and ethical use of AI in healthcare.

% Optionally include supplemental material (complete proofs, additional experiments and plots) in appendix.
% All such materials \textbf{SHOULD be included in the main submission.}

\end{document}